\documentclass[letterpaper, 10 pt, journal, twoside]{IEEEtran}
\IEEEoverridecommandlockouts                              % 
\usepackage[colorlinks,linkcolor=red,anchorcolor=blue,citecolor=green]{hyperref}

\usepackage{cite}
\usepackage{amssymb,amsfonts}
\usepackage{amsmath}
\usepackage{multirow}
\usepackage{graphicx}
\usepackage{textcomp}
\usepackage{xcolor}
\usepackage{times}
\usepackage{soul}
\usepackage{url}
\usepackage{graphicx}
\usepackage{subfigure}
\usepackage{float}
\usepackage{bm}
\usepackage{booktabs}
\usepackage{graphicx}
\usepackage{booktabs}
\usepackage[ruled]{algorithm2e} 
\usepackage{algorithmicx}
\usepackage{algpseudocode} 
\usepackage[ngerman]{babel}
\usepackage{paralist, tabularx}
\usepackage{hyperref}
\usepackage{subfigure}
\usepackage{color,xcolor}
\begin{document}

\title{A Robotic Visual Grasping Design: Rethinking Convolution Neural Network with High-Resolutions }
%


%\author{Michael~Shell,~\IEEEmembership{Member,~IEEE,}
    %        John~Doe,~\IEEEmembership{Fellow,~OSA,}
    %        and~Jane~Doe,~\IEEEmembership{Life~Fellow,~IEEE}% <-this % stops a space

\author{Zhangli Zhou\textsuperscript{1} , Shaochen Wang\textsuperscript{1}, Ziyang Chen\textsuperscript{1}, Mingyu Cai\textsuperscript{2}, and Zhen Kan\textsuperscript{1} \thanks{$^{1}$Z. Zhou, S. Wang, Z. Chen, and Z. Kan (corresponding author) are with the Department of Automation, University of Science and Technology of China, Hefei, China. Email: \tt\footnotesize zzl1215@mail.ustc.edu.cn, samwang@mail.ustc.edu.cn, chenziyang19981220@126.com, zkan@ustc.edu.cn}
        % , 230026. e-mail: \url{zzl1215@mail.ustc.edu.cn}, \url{samwang@mail.ustc.edu.cn}, \url{chenziyang19981220@126.com}, \url{whyy@mail.ustc.edu.cn}, \url{zkan@ustc.edu.cn.}
\thanks{$^{2}$M. Cai is with the Department of Mechanical Engineering and Mechanics, Lehigh University, Bethlehem, PA, USA. Email: \tt\footnotesize mingyu-cai@lehigh.edu}
    }
    
\maketitle
\thispagestyle{empty}
\pagestyle{empty}




\begin{abstract}

High-resolution representations are important for vision-based robotic grasping problems. Existing works generally encode the input images into low-resolution representations via sub-networks and then recover high-resolution representations. This will lose spatial information, and errors introduced by the decoder will be more serious when multiple types of objects are considered or objects are far away from the camera.
To address these issues, we revisit the design paradigm of CNN for robotic perception tasks. We demonstrate that using parallel branches as opposed to serial stacked convolutional layers will be a more powerful design for robotic visual grasping tasks.
In particular, guidelines of neural network design are provided for robotic perception tasks, e.g., high-resolution representation and lightweight design, which 
respond to the challenges in different manipulation scenarios. We then develop a novel grasping visual architecture referred to as HRG-Net, a parallel-branch structure that always maintains a high-resolution representation and repeatedly exchanges information
across resolutions. Extensive experiments validate that these two designs can effectively enhance the accuracy of visual-based grasping and accelerate network training. We show a series of comparative experiments in real physical environments at \url{https://youtu.be/Jhlsp-xzHFY}.

% We demonstrate that using parallel branches as opposed to serial stacked convolutional layers will be a more powerful design for robotic visual grasping tasks, which can also be applied to general vision-based robotic applications.
% Such problems are exacerbated when multiple types of objects are considered or objects are far away from the camera.
% Since objects occupy fewer pixels, the errors introduced by the decoder will appear more obvious.
\end{abstract}

		
		% \fi
\IEEEpeerreviewmaketitle

\section{INTRODUCTION}

Since the world's first robot in 1960s, robots have been playing more and more roles in intelligent manufacturing, domestic services, etc. Pick and place is one of the basic skills in robot manipulation. Choosing appropriate grasping position and pose based on visual perception is crucial for such manipulations. 

As a variant of convolutional neural networks (CNN) \cite{DBLP:conf/icra/MorrisonCL19, lenz2015deep, kumra2020antipodal, ainetter2021end,kumra2017robotic,Redmon2015,wang2016robot}, deep CNNs have become the engine of visual perception and recognition. Considerable progress including LeNet \cite{lecun1998gradient},  AlexNet\cite{krizhevsky2017imagenet}, VGGNet\cite{wang2015places205}, and ResNet\cite{he2016deep}  advances and improves the perceptual capacity of the models. Although the aforementioned models were initially designed for natural image classification,
a lot of subsequent works in the robotic community adopt the architecture used in these models as their backbone. However, the robotic visual perception tasks are quite different from the classical natural image processing tasks. Typical applications in industrial robots, such as assembling and drilling, not only require recognition of different objects but also manipulation of them in various scenarios. Unlike general visual classification that only needs semantic representation of objects, robotic perception tasks more rely on accurate spatial information of the objects. This is because the model often encounters objects that are not seen in the dataset of training.
\begin{figure}
    \centering
    \includegraphics[width=0.45\textwidth]{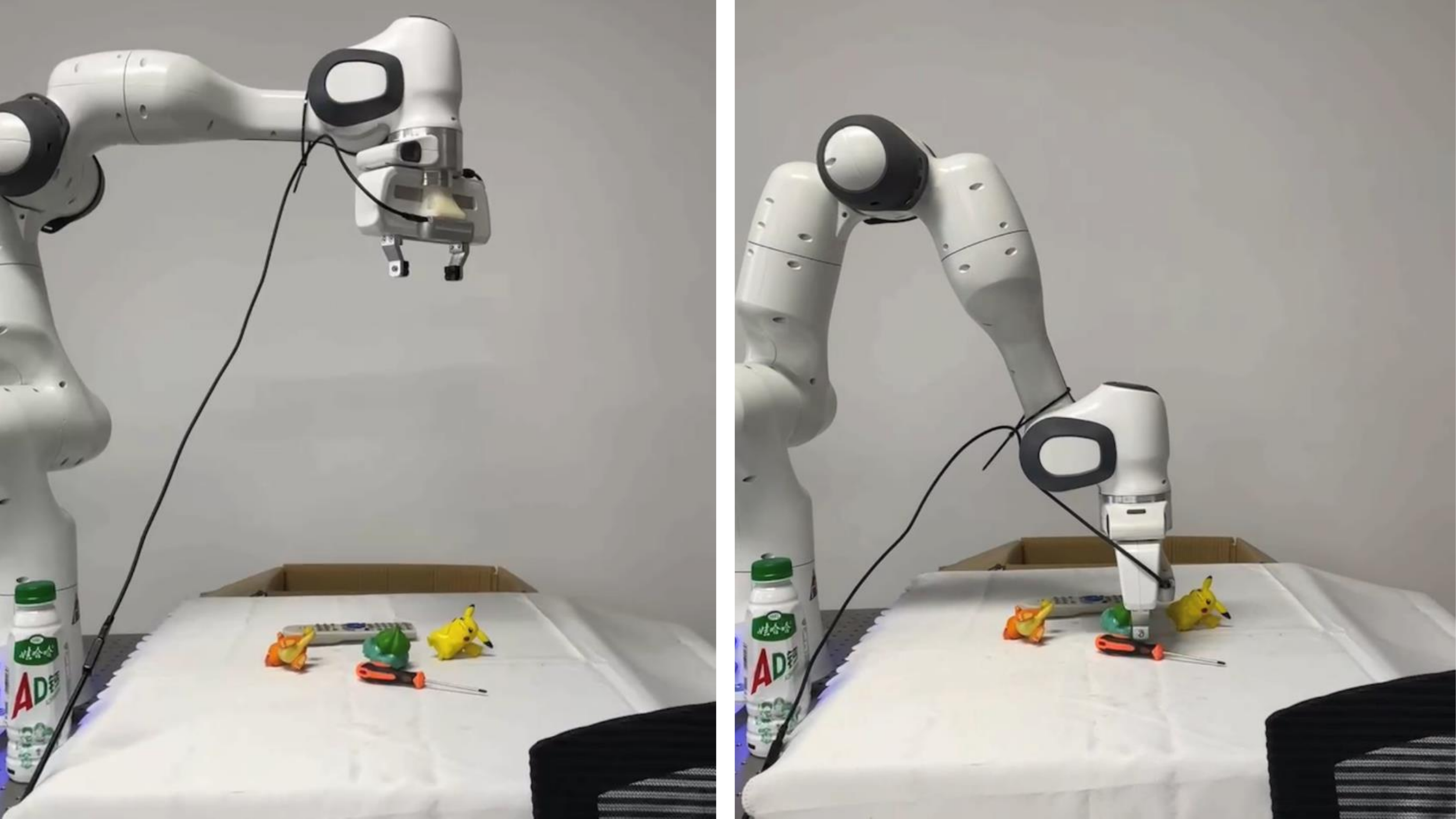} 
    \caption{Conventional encoder-decoder architectures are not spatially accurate enough and may fail to grasp stacked objects.}	
    \label{failed}
\end{figure}

In this work, we focus on retaining the details of the representation. Previous approaches mainly used encoder and decoder structures, and U-Net \cite{Ronneberger2015} like networks can extract features well and contain rich semantics. However, spatial information can lose when recovering high-resolution representations from the encoded low-resolution representations, resulting in inappropriate predictions and failure of grasping tasks.
As shown in Fig.~\ref{failed}, the workspace is cluttered with various objects. When approaching to the target object, the claw touches an adjacent object and thus fails in grasping the target object. To address this issue, we present a novel neural network design paradigm for robotic visual grasping as shown in Fig.~\ref{network}. It is a parallel-branch structure that maintains a high-resolution representation and repeatedly exchanges information across resolutions. The model not only reaches the state-of-the-art results on several mainstream datasets ($99.5\%$ in Cornell\footnote{ Cornell
dataset: \url{http://pr.cs.cornell.edu/
grasping/rect_data/data.php.}}\cite{Lenz2015} and $96.5\%$ in Jacquard \cite{depierre2018jacquard}),  but also perform well in various grasping-related applications in physical environments. To unravel the role of components in our model, the ablation experiments are performed. It is also found that high resolution is crucial for robotic perception tasks.
Our insight is that the model should more focus on the configuration information, e.g., object position, shape, etc, so that stronger perception can be obtained to facilitate robotic grasping. 
The main contribution of this paper can be summarised as follows:
\begin{compactitem}[$\bullet$]
    % \item  We explore the high-resolution feature sensing in a visual robotic grasping task. To the best of our knowledge, it is one of the first attempts to explore the high-resolution feature map for  robotic visual grasping.

    \item To the best of our knowledge, it is one of the first attempts to explore the high-resolution feature map for improved perception quality in robotic visual grasping. 
    
    \item We develop an online closed-loop robotic grasping framework that dynamically adjusts the motion of the end-effector based on real-time visual detection.

    % \item We deployed our model in a real physical environment and tested the performance of our method against other methods for multi-object grasping, in addition to applying the method to a continuous block building task to demonstrate the robustness of our method.

    \item  Extensive comparison results and physical experiments show the advantages of maintaining high-resolution features in the robotic perception framework.
    
    % \item Exhaustive experiments are conducted to show the advantages of maintaining a high-resolution feature in the robotic perception framework. The results show that our model achieves the best current results on the dataset. Various ablation experiments also confirm the effectiveness of maintain high-resolution features.

\end{compactitem}

% \section{Related Work}

\textbf{Related Work:\text{ }} Robot grasping in general consists of three main components: grasping detection, trajectory planning, and motion control. Grasping detection is to generate an executable grasping configuration that maximizes the success rate of grasping tasks. These methods broadly fall into two categories: geometry-driven methods and data-driven methods. Geometry-driven approaches generally obtain a grasp by analyzing and calculating the geometry and kinematics of the physical model. A major limitation is that geometry-driven approaches typically assume a known physical model, which is not often available or accurate in practice. 
Data-driven approaches are mostly based on learning methods and usually require a large number of human-labeled samples.
Prior classical methods used handcraft feature engineering for vision-based grasping \cite{ramisa2014learning}. With the flourishing of deep learning, end-to-end training approaches are being gradually used instead.
Learning-based methods show their potential in robotic grasping. 
The method of representing the grasp pose by a grasping rectangle is proposed by \cite{jiang2011}.  The work \cite{lenz2015deep} proposes a two-state solution, where the first step is to generate a large number of grasp rectangles and the second stage evaluates the quality of each rectangle and chooses the best one among them.
Recent works \cite{lenz2015deep,DBLP:journals/ral/ShaoFJNLSOKB20,zhang2019roi,morrison2020learning,cao2021lightweight} utilize deep neural networks to complete the grasping detection task, in which convolution neural networks are applied to perform vision-based grasping from raw sensor inputs. Afterwards, region proposal network (RPN) \cite{ren2015faster} is introduced into grasping planning by \cite{DBLP:conf/icarm/LuoTJPX20}.  The RPN is primarily designed to regress the bounding box of objects and avoid sampling candidate boxes, reducing the time complexity. On the other hand, larger and deeper models \cite{DBLP:conf/icra/PintoG16, kumra2017robotic, wang2022transformer} are also used to improve the performance of grasp detection.  Kumra et al. \cite{kumra2017robotic} present a 50 layers ResNet and achieve high accuracy.  So far, the majority of these algorithms are based on encoder and decoder architecture. However, extensive experiments show that this method is very sensitive to the distance between the object and the camera and the clutter of the workspace on the grasping task, which often leads to failure of the grasping task due to poor prediction of position and pose.
		
 %To distinguish from methods mentioned  above, we scale the breakthrough in natural language processing (e.g., BERT, GPT), using unsupervised pre-training to obtain a better representation of words.
%Our method incorporates self-supervised representation learning, reducing the need for annotated samples, and  measures the similarity of objects under multi viewpoints to achieve a robust representation. 
%It is also different from previous methods that exploit high-capacity models to improve grasping detection performance\cite{DBLP:conf/icra/PintoG16}, \cite{kumra2017robotic}.
%When the models like deep residual networks are deployed on real robots, the real-time performance would be harmed due to its slow inference speed. In contrast, the models employed  self-supervised representation learning can improve the system performance without significantly adding model parameters.

		%High-resolution was first developed for pose estimation.
		
		%High-resolution representations have been shown to be effective for a variety of vision tasks such as human pose estimation, segmentation, and object detection.
		
		%There are two approaches to achieve the high-resolution representation. One is to recover from a low-resolution feature map, while the other is to maintain the high-resolution directly via high-resolution convolutions.
 \section{Backgroud and Problem Formulation}

\subsection{Grasp Representation}
Our network output consists of three feature maps $\mathcal{Q}$, $\Phi$, and $\mathcal{W}$, which are of the same size as the input raw depth image as shown in Fig. \ref{network}. we introduce $\mathcal{Q}$ to better reflect the likelihood of a successful grasp when each pixel in the image is used as the center of the grasp rectangle, the parameters in the grasping quality score $\mathcal{Q}$ take values between $0$ and $1$.  When the value of a place in the quality score $\mathcal{Q}$ is closer to $1$, it means that the probability of a successful grasp at that location is higher. Subsequently, the best grasping point can be obtained by retrieving the grasping quality score $g^*=\arg \max_\mathcal{Q} g_i $. $\Phi$ is an image that describes the grasp angle to be executed
at each point $p$. Since the antipodal grasp is
symmetrical around $\pm\frac{\pi}{2}$ radians, the angles are given
in the range $\left[-\frac{\pi}{2},\frac{\pi}{2}\right]$. $\mathcal{W}$
is an image that describes the gripper width to be executed
at each point $p$. To allow for depth invariance,
the variable $z$ is in the range of {[}0, 150{]} pixels, which can be converted
to a physical measurement using the depth camera parameters and the measured
depth. In this work, we adopt a five-dimensional dynamic evolution for the parallel gripper to represent the grasp representation. The corresponding gripping configuration includes grasping center point $p = (x,y)$, rotation angle $\theta$, the width of robot manipulation gripping jaws $w$ in $p$ and height $z$ of end-effector. A complete grasping rectangle $g$ can be defined as 
 \begin{equation}
     g=(x,y,\theta, w,z), (x,y) \in \mathcal{W}, \theta \in \Phi, w \in \mathcal{W}
 \end{equation}
To calculate the predicted position and pose, we just need to search for the maximum value in $\mathcal{Q}$ to get the coordinates of the best grasping point $(x^*,y^*) = \arg \max_\mathcal{Q}(x,y)$. Then, based on the coordinates of this point $p = (x^*,y^*)$, $w$ and $\phi$ can be obtained directly in the width map $W$ and angle map $\Phi$, so that the predicted grip position and pose are completely known. 
The distance $z$ of the jaws from the object is measured by the depth camera, which takes into account the size of the jaws and the placement of the camera.

 \subsection{Problem Formulation}
Given some labeled dataset (Cornell or Jacquard, etc.), select $N$ samples $(x_1, x_2, ..., x_N), x_i \in \mathbb{R}^{H \times W}$, $H$ and $W$ are the height and width of the image, respectively. The objective of visual net is about how to construct a neural network $\mathcal{F}$ with parameters $\Theta$ to ensure that the loss function is minimized after $\Theta$ certain number of updates. The loss function is:
 \begin{equation}
     L(\Theta)=\frac{1}{2 N} \sum_i^N\left(\mathcal{F}\left(x_i, \Theta\right)-y_i\right)^2
 \end{equation}
where $y_i$ is label corresponding image $x_i$. 

After obtaining the perception module, the next objective is to integrate it into the motion planning to online adjust manipulation trajectories to accomplish grasping tasks. In the following section, we'll first show the current limitations of existing works and then propose our solution.

% {\color{red} In a workspace containing multiple objects, there may be adhesions or overlaps between multiple objects. Initially, the camera is far away from the objects, and the position and pose g predicted at the beginning may not be very accurate. How to achieve more reasonable motion planning and prediction of grasping position and pose based on the information obtained by calling the model.}

\section{HRG-Net Based grasping framework}
\begin{figure}
\centering
\includegraphics[width=0.35\textwidth]{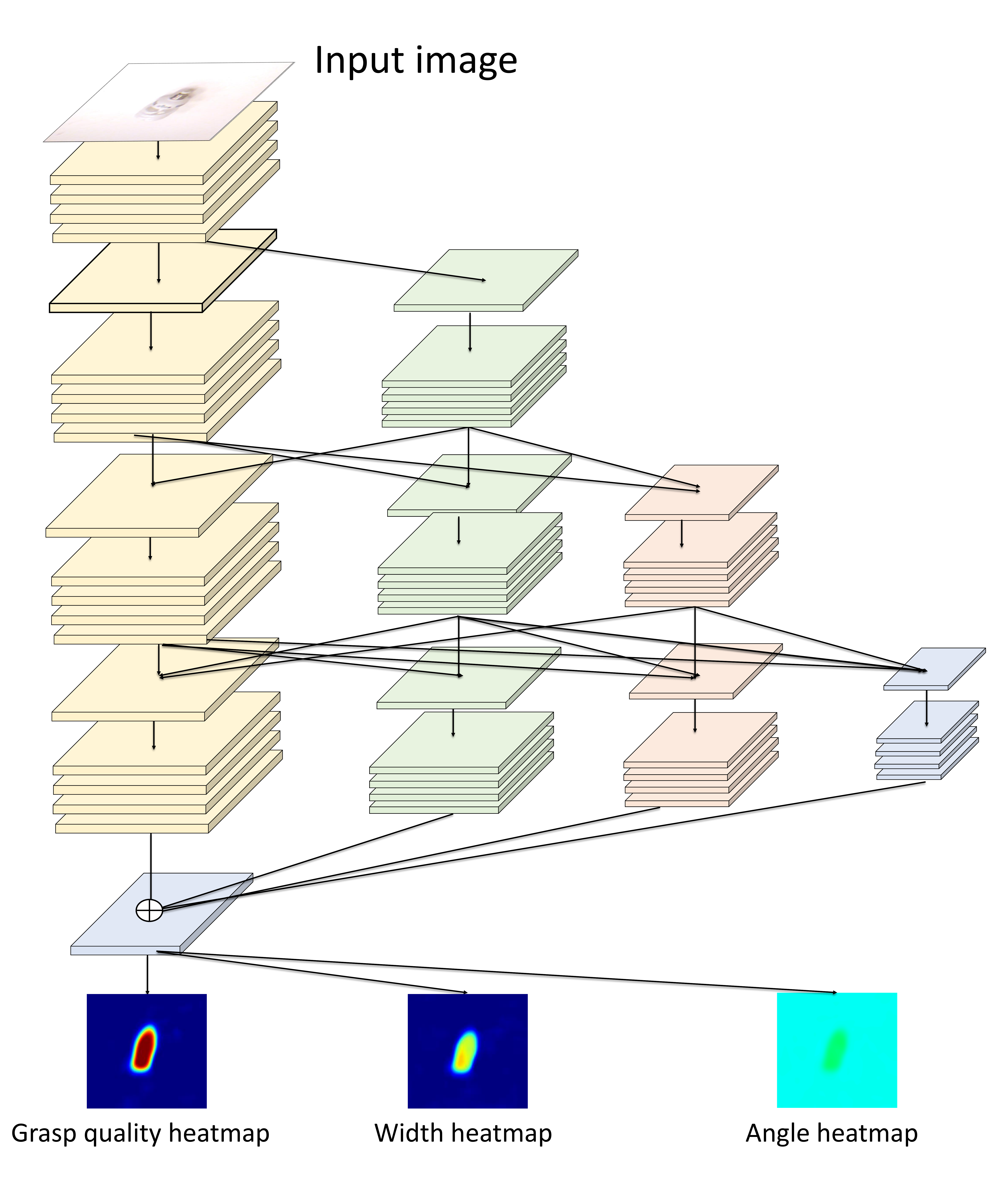} 
\caption{The network architecture of HRG-Net, unlike traditional CNNs, maintains a relatively high-resolution feature map throughout the network.}
\label{network}
\end{figure}

%  {\color{blue} Mingyu: Be careful about this section (three subsections) and improve the descriptions concretely and technically.}

\subsection{Motivation}

Feature extraction plays a crucial role in robotic visual grasping.  One early approach~\cite{ramisa2014learning} applied hand-crafted features to extract the grasping candidates, and then used methods of deep learning to stack convolutional layers and build abstract feature representation. Unlike the general supervised vision tasks that often demand refined semantic information such as object category features, grasping models pay more attention to fine-grained geometric features such as series of edges and the shape of an object to facilitate grasping. Existing mainstream works~\cite{Morrison2018,DBLP:journals/ral/ShaoFJNLSOKB20} generally encode the input image into a low-resolution representation through a subnetwork and then decode the high-resolution representation from it. But in practice, we realized that such a common visual strategy can not help the robot accurately predict gripping configurations when the object is far away from the camera or multiple objects are stacked together. As analyzed in Mask-RCNN~\cite{he2017mask}, this downside is caused by upsampling the feature map that is too small. Inspired by \cite{sun2019deep, wang2020deep}, we aim at maintaining a high-resolution feature map throughout the overall visual process to avoid loss of spatial accuracy. Such an application is not trivial since we need to consider robotic practical situations and bridge gap of sim-to-real.

% Similar as human grasping, the shape of objects instead of only semantic information should be integrated with vision systems.
% input image as a low-resolution representation through a subnetwork that is formed by connecting high-to-low resolution convolutions in series, and then recover the high-resolution representation from the encoded low-resolution representation. But in practice we found when the object is far away from the camera, or when multiple objects are stacked together, the robotic arms often do not accurately predict gripping position and attitude. We are inspired by ROI align in Mask-RCNN \cite{he2017mask} and suspect that it is due to the lack of accuracy of the predicted position caused by upsampling after the feature map is too small, so we need to maintain a high-resolution feature map throughout the processing to avoid loss of spatial accuracy.

\subsection{HRG-Net Architecture}

We present a design that differs from prior encoder-decoder architectures\cite{Morrison2018, Lenz2015, kumra2017robotic} for robotic perception. 
As shown in Fig.~\ref{network}, The entire network framework can be divided into $4$ stages. Each stage consists of parallel stacked block with different feature resolutions, and each block is a residual block. Information interaction between different blocks is carried out via FuseLayer, where features are transitioned to the high-resolution branch by up-sampling operation and vice versa by down-sampling to the low-resolution branch.
Concretely, the model utilizes a $3\times 3$ convolution kernel with stride $2$ to decrease the feature map resolution and a bilinear interpolation to perform up-sampling.
The high-resolution representation fusion is obtained by parallelly mixing different resolution convolutional layers. Consequently,
our approach can preserve the high-resolution representation by parallelly connecting high-to-low resolution convolutions, and iteratively performing fusion operations between parallel blocks.

 The main characteristics of our model are  \romannumeral1) parallel connections from high-resolution to low-resolution throughout all phases of the model, and \romannumeral2) exchanges of information across different resolutions to enrich semantic information. Technically, the network first starts with a convolutional stem block, gradually stacks convolutional blocks with different resolutions, and connects them in parallel.  In general, the features learned by HRG-Net are both semantically and spatially strong. This is because the convolutional blocks of different resolutions are linked in parallel rather than serially, which is more beneficial for learning accurate spatial position information. And our model consistently maintains a high-resolution feature representation, instead of shrinking the feature maps as in traditional encoders. 
In addition, there is an ongoing fusion of information among the different branches, making a wealth of information at the semantic level. 

% The entire network framework can be divided into $4$ stages. Each stage consists of parallel stacked blocks with different feature resolutions, and each block is a residual block. Information interaction between different branches is carried out via FuseLayer, where features are transitioned to the high-resolution branch by up-sampling operation and vice versa by down-sampling to the low-resolution branch.
% Concretely, the model utilizes a $3\times 3$ convolution kernel with stride $2$ to decrease the feature map resolution and a bilinear interpolation to perform up-sampling.
% The high-resolution representation fusion is obtained by parallelly mixing different resolution convolutional layers. As a result,
% our approach can preserve the high-resolution representation by parallelly connecting high-to-low resolution convolutions, and iteratively performing fusion operations between parallel branches.
% The resulting representation is rich in terms of both semantic and spatial information.
% \textcolor{blue}{Finally, we apply a pixel-level grasp prediction~\cite{loshchilov2018decoupled}.}

% The resulting representation is not only semantically rich but also precise in terms of spatial position.
% To allow for more attention to detail, we apply a pixel-level grasp prediction \cite{loshchilov2018decoupled}.

\subsection{Vision-based Trajectory Planning}

% In this section, let's detail the trajectory planning algorithm we use. First of all, it must be noted that our entire gripping process is a dynamic， closed-loop and real-time, where the robotic arm end-effector uses the camera to capture images as input to the network during the process from the highest point $z_{max}$ to the lowest point $z_{min}$. First, the choice of camera viewpoint plays an important role in the quality of visual grasping detection. In this work, we apply active perception techniques \cite{morrison2019multi} to calculate the next best viewing angle in real-time with an eye-in-hand camera as the robot reaches $z_{min}$. This gives us a trajectory $((p_0, p_1, ...), p_i = (x,y,z) \in \mathbb{R}^3, (p_i[2] - p_j[2]) \times (i - j) \textgreater 0)$ of decreasing height.

Initial visual predictions with far distance to complex objects is crucial for the overall grasping task.
This section introduces a trajectory planning framework by leveraging HRG-Net that has better performance regarding this point. 
First, it is worth noting that our entire gripping is a closed-loop process that adjusts trajectories of the manipulation's end-effector in real-time using visual information from the camera as feedback. It online captures input images as the end-effector moving close to the object vertically. Consequently, the choice of camera viewpoint plays an important role in the quality of visual detection.  We apply an active perception technique \cite{morrison2019multi} to calculate the next best viewing angle in real time with an eye-in-hand camera.
Before completing the grasping task, let $z_{max}$, $z_{obj}$ denote the heights of the initial end-effector and the predicted height of the object. We can have a trajectory $((p_0, p_1, ...), p_i = (x,y,z) \in \mathbb{R}^3, (p_i[2] - p_j[2]) \times (i - j) \textgreater 0)$ with decreasing heights (from $z_{max}$ to $z_{obj}$).
During this process, the HRG-Net model is called when a new viewpoint is reached. Thus, we can obtain a series of predictions $((g_0, g_1, ...), g_i = \mathcal{F}((camera(p_i),\Theta)$ from the trajectory points $((p_0, p_1, ...)$, where $camera(p_i)$ is the image captured by the camera at $p_i$. Based on the direction of decreasing entropy of the grasping quality map $Q$, the robot arm then executes the horizontal trajectory planning and descends in height at a specific rate. Our trajectory planning method can effectively avoid clutter or occlusion situations owing to multi-view in trajectory.

% {\color{blue} (Mingyu: confusing descriptions:) But the trouble is that the camera moves with the robot arm, which will cause the spatial error brought by the decoder in the ordinary encoder-decoder architecture to be amplified again, directly leading to a very inaccurate optimal gripping position calculated after synthesizing the whole trajectory.}

% {\color{blue} (Mingyu: English problems:) Fig. \ref{sim} shows a partial screenshot of the full grasping process, multiple objects adhering to each other on the workspace, the starting point of the whole trajectory planning process is the center of the workspace is placed at $z_{max} = 60cm$. }

\begin{figure}[htbp]
\centering
\subfigure[Start point for trajectory planning]{
    \includegraphics[width=3cm]{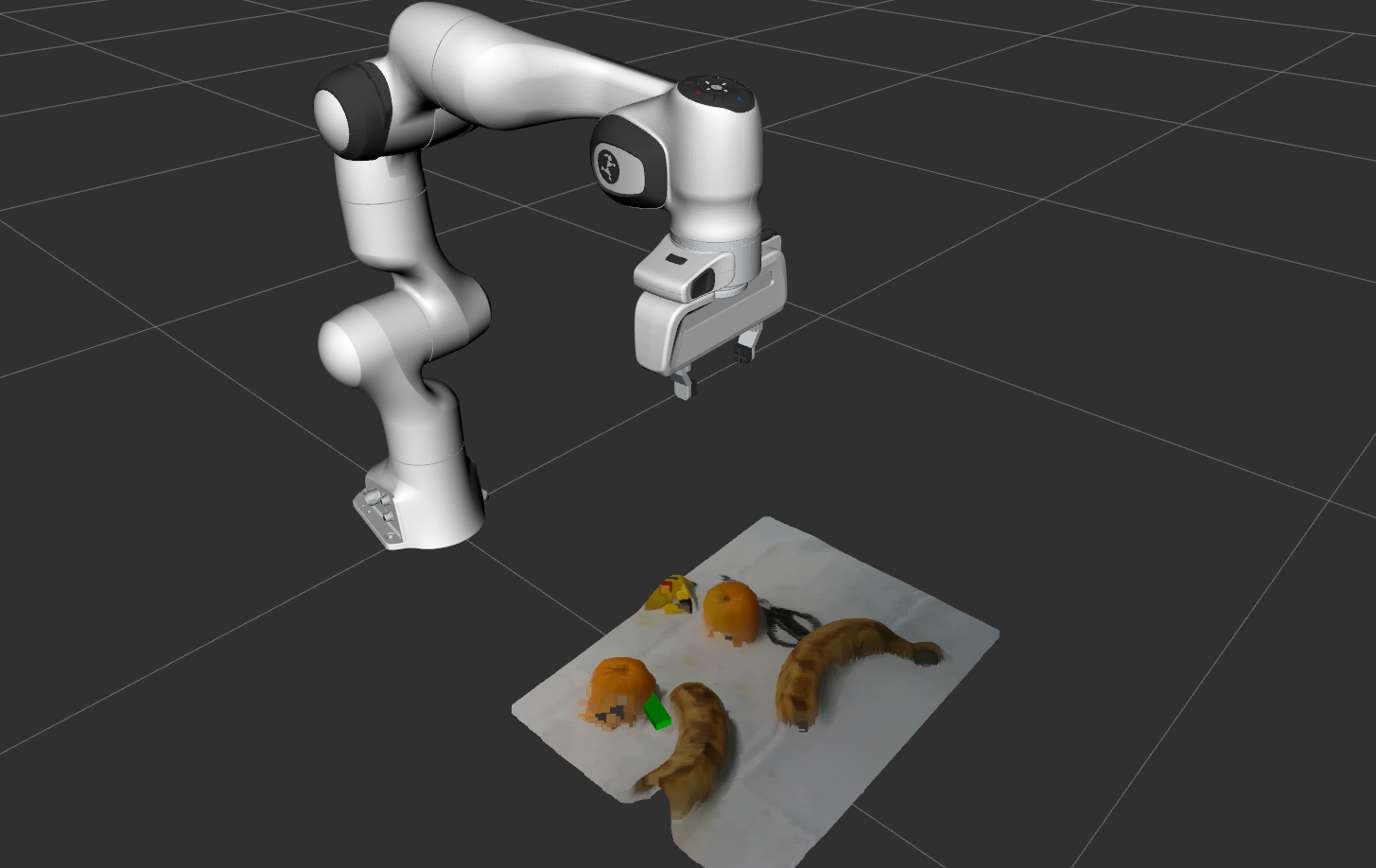}
    %\caption{fig1}
}
\quad
\subfigure[Intermediate point of the trajectory planning]{
    \includegraphics[width=3cm]{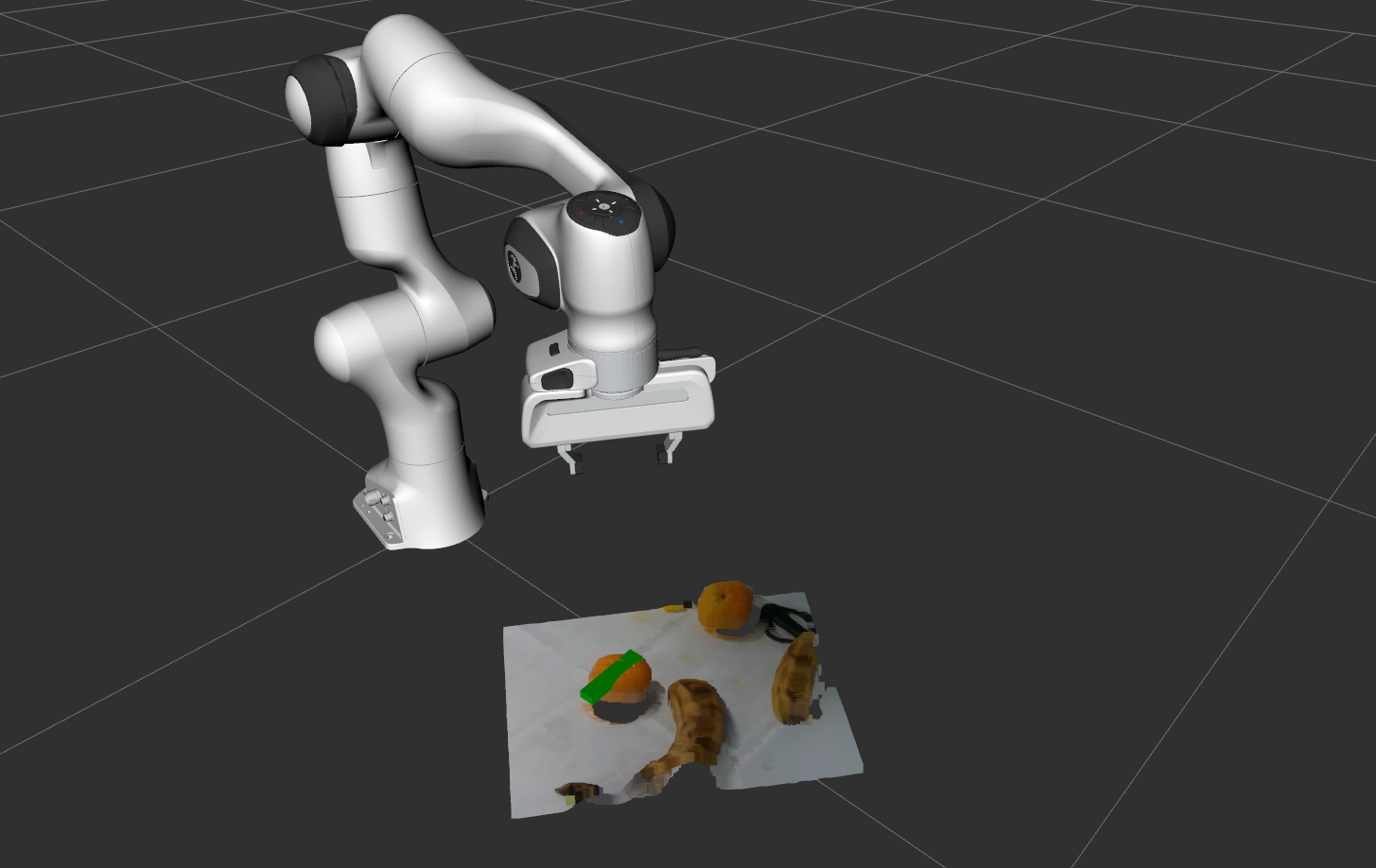}
}
\quad
\subfigure[Intermediate point of the trajectory planning]{
    \includegraphics[width=3cm]{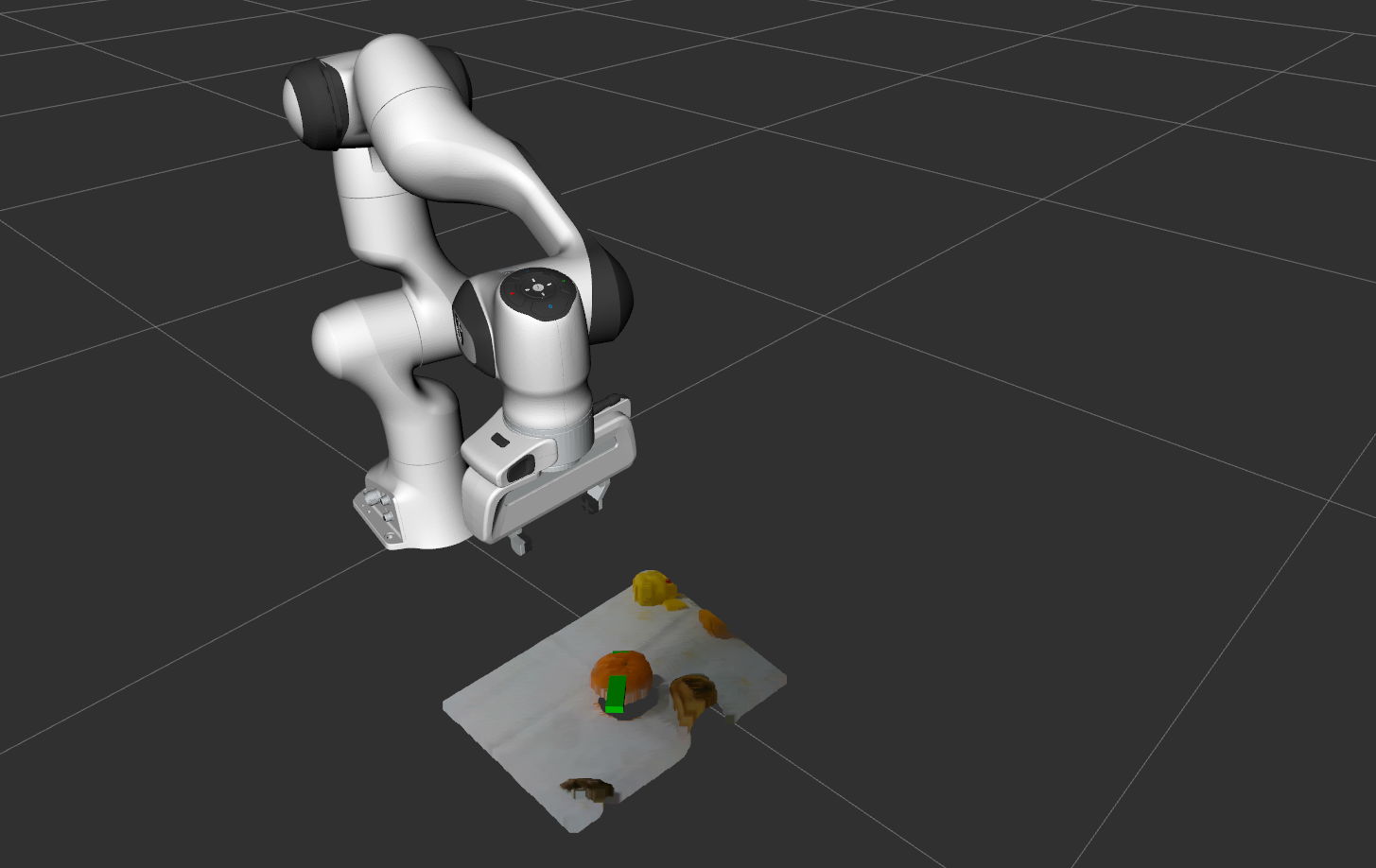}
}
\quad
\subfigure[End point of the trajectory planning]{
    \includegraphics[width=3cm]{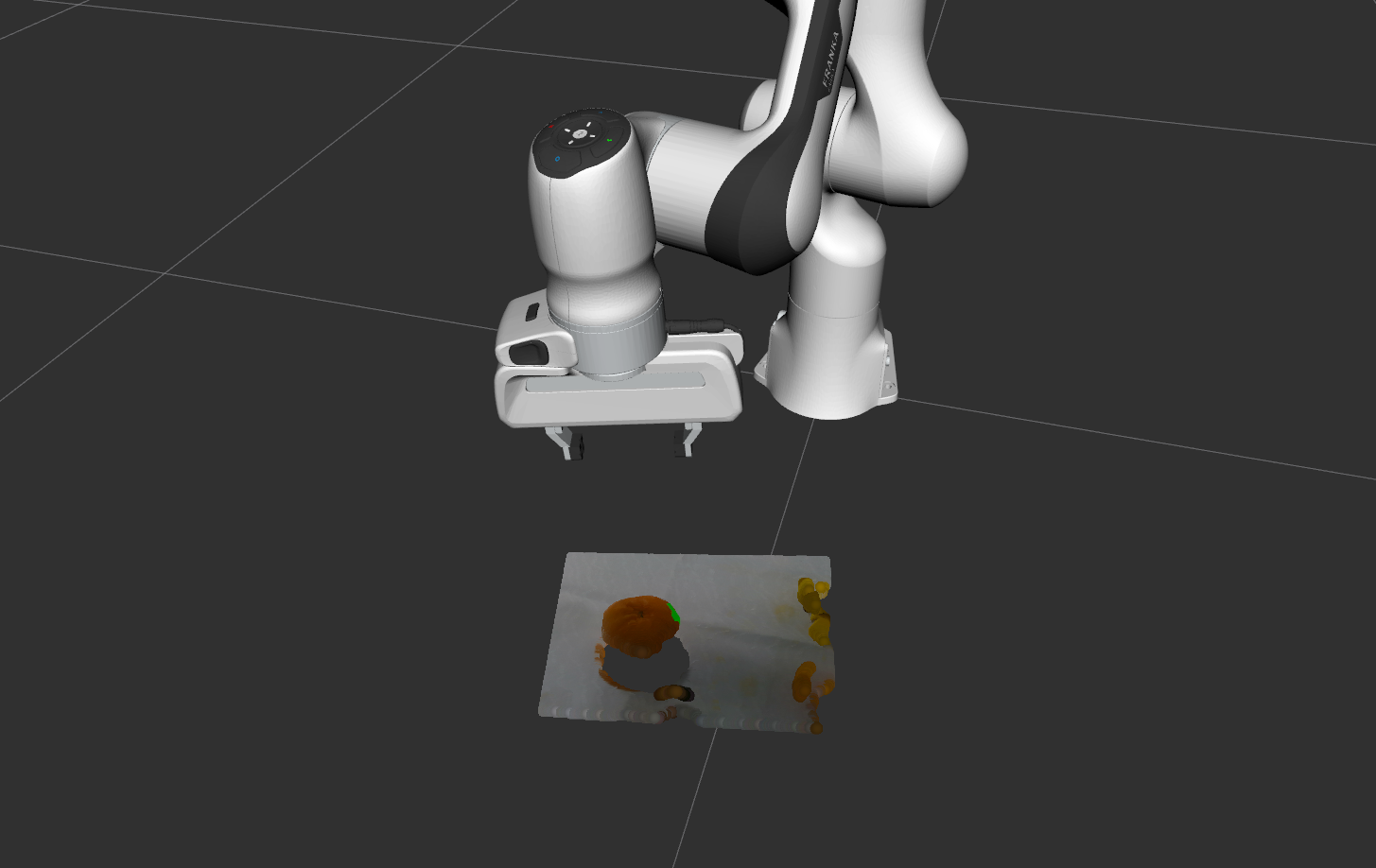}
}

\caption{ These snapshots show that multiple calls of HRG-Net can improve grasping prediction performance. (a)-(d) reflect the multi-visual process of grasping an orange. The objects observed by the camera are shown using the point cloud in the simulation, and the green rectangular blocks are visualizations of the predicted grasping position and pose.  }
\label{sim}
\end{figure}

As a running example in Fig.~\ref{sim},
we can see that HRG-Net initially gives an inaccurate prediction (green solid rectangle) in Fig.~\ref{sim} (a).
At this point, the direct grasping based on offline motion planning may lead to a collision with the banana resulting the failure. By applying our method, as the end of the robot arm gradually approaches the workspace shown in Fig.~\ref{sim} (b)-(c), the camera's perspective decreases and the prediction will be more precise.
Finally, when the height of the robotic manipulation's end-effector does not reach the object i.e., $z \geq z_{obj}$ in Fig.~\ref{sim} (d), HRG-Net gives a most reasonable prediction from previous perception information.  
 
%  {\color{blue} Mingyu: The logic of this section needs to be improved. First, it needs to clearly descriptions for the approach. Then, we can show an running example for overall explanation. The center of the method is to focus on the close-loop structure and adjust trajectory planning online using visual information.}
		
\section{Experiment}

In this section, extensive experiments are carried out to answer why
the robot perception models need to keep high-resolution representations.
To demonstrate our metrics, experiment results including comparisons with baselines~ \cite{Lenz2015,Morrison2018,kumra2017robotic} that use sequential stacked convolutional layers.

% demonstrate that the developed high-resolution
% the model outperforms existing baselines using sequential stacked convolutional layers \cite{Lenz2015,Morrison2018,kumra2017robotic}. 

% Why do the robot perception models need to keep high-resolution representation? Empirical evidence is provided in this section to show that the high-resolution model enables more concentration on details, and thus the final results are superior to other counterparts with sequential stacked convolutional layers.
	
\subsection{System Setup}

The 7 DoF Franka robot with 2-finger gripper is used to perform grasping. The RealSense D435 RGB-D camera is mounted on the end-effector of the robot for accurate predictions of gripping position and pose. Ubuntu realtime kernel is used to control the robot manipulation and an additional computer with a 3.6Ghz Intel i7-9700k CPU and Geforce RTX 2080 Super GPU is used for the network deployment and computing. The system is built on ROS.

\subsection{Implementation Details}

For our high-resolution grasp model, we utilize AdamW  optimizer\cite{DBLP:conf/iclr/LoshchilovH19} with an initial learning rate of $1e-4$ and weight decay of $5e-2$. The input resolution is croped to $224 \times 224$ and the input batch is $I(N,C,H,W)$.  The momentum of BN layer is set to $0.1$. The batch size is set as $32$ and the epoch is set as $50$.
		
\subsection{Experimental and Analysis in Datasets}

\begin{figure}[htbp]
\centering
\subfigure[HRG-Net in Cornell]{
    \includegraphics[width=3.5cm]{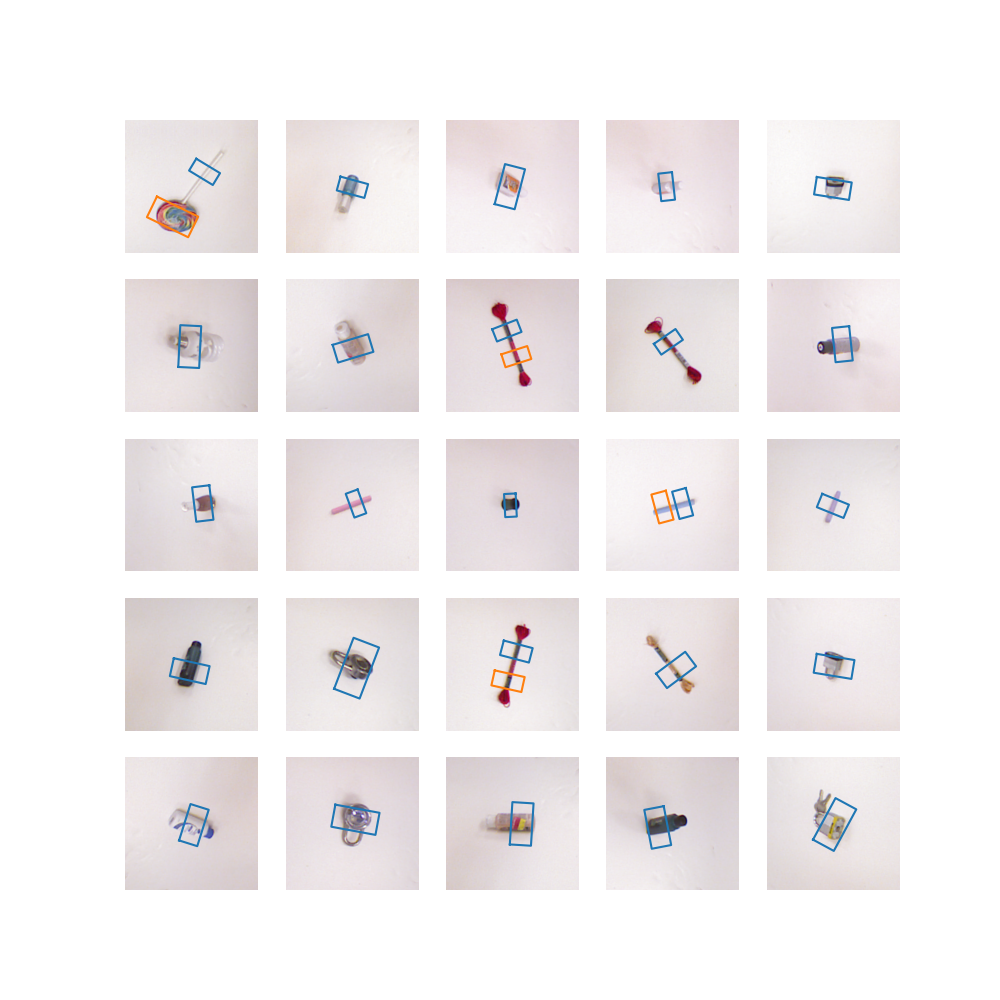}
    %\caption{fig1}
}
\quad
\subfigure[HGR-Net in Jacquard]{
    \includegraphics[width=3.5cm]{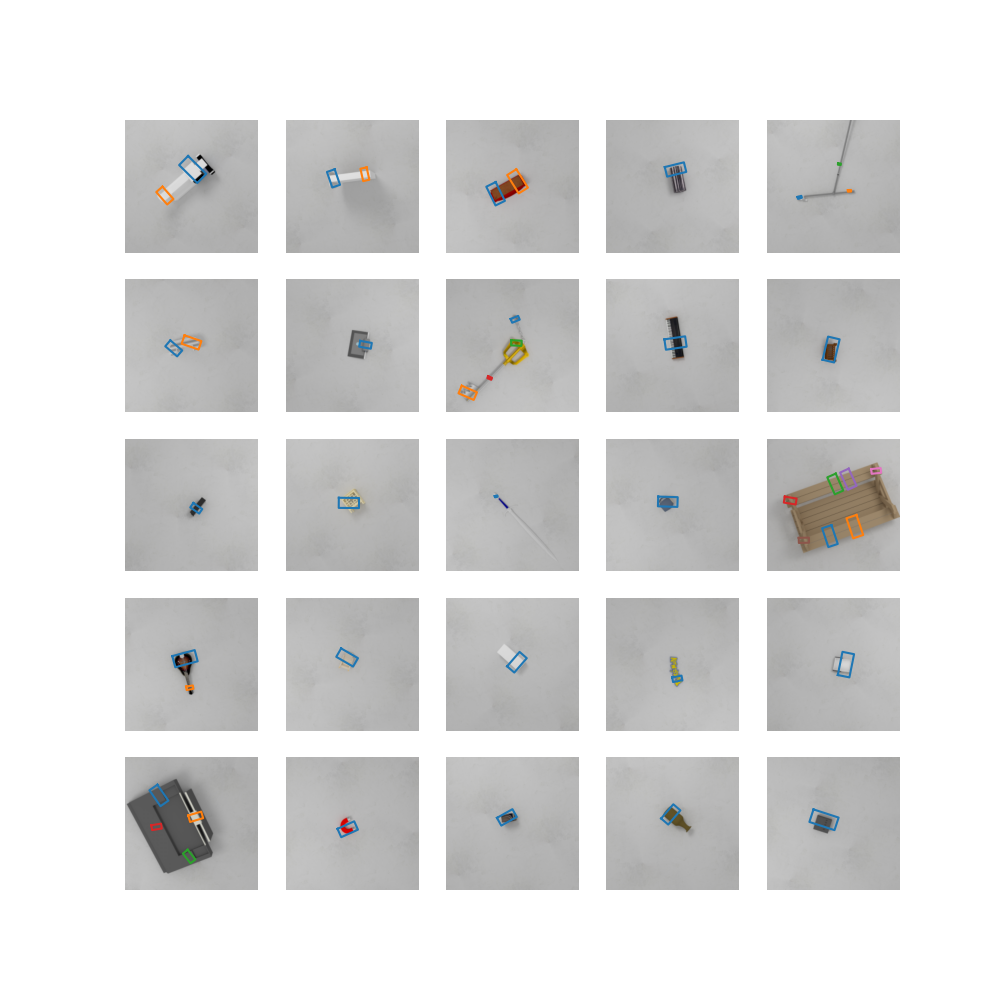}
}
\quad
\subfigure[HRG-Net in unseen multiobject]{
    \includegraphics[width=3.5cm]{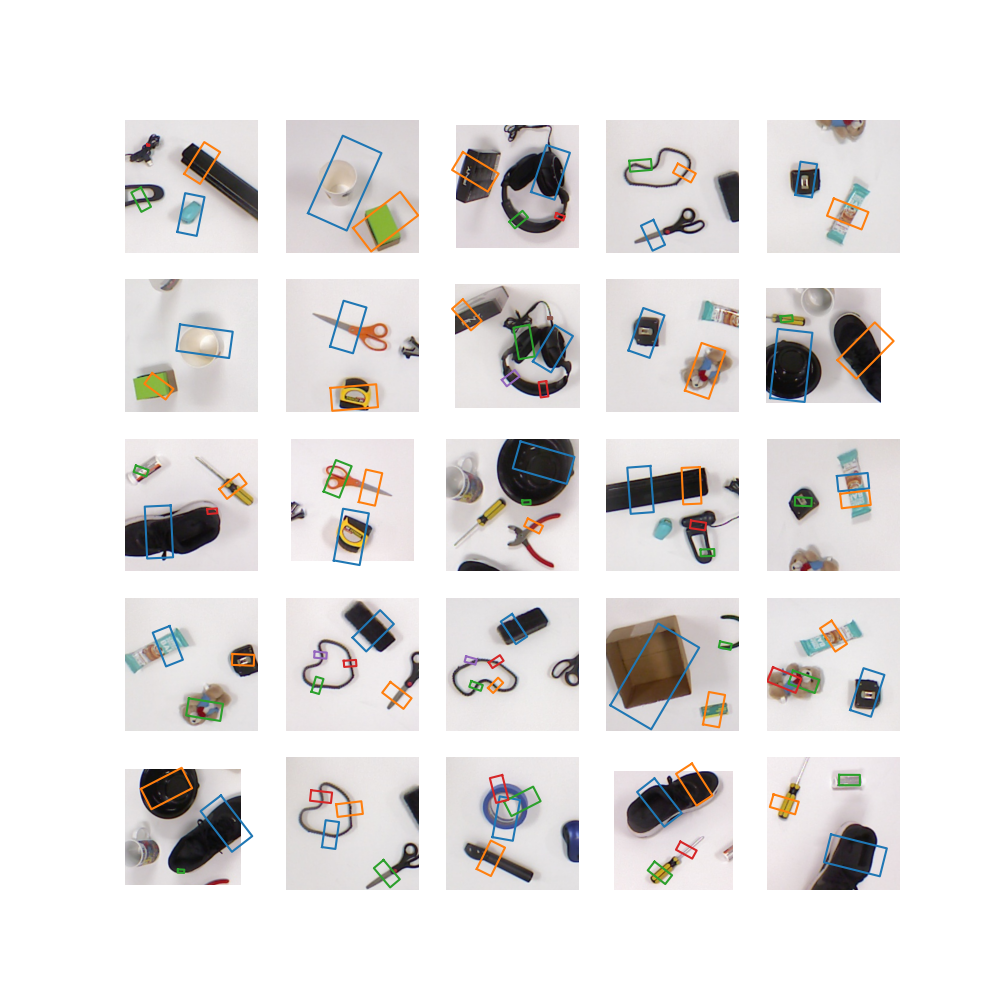}
}
\quad
\subfigure[Ablation study in Cornell]{
    \includegraphics[width=3.5cm]{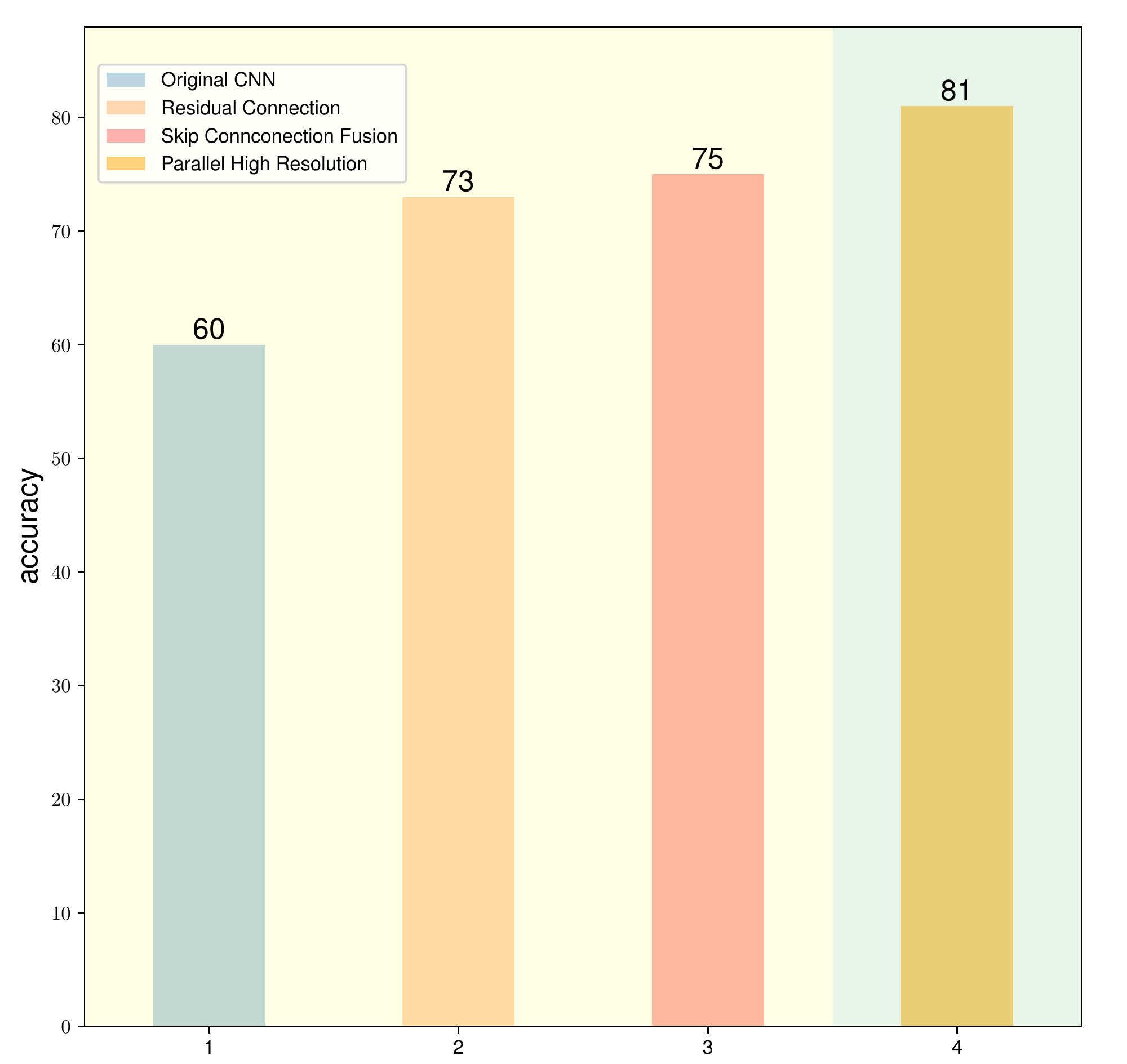}
    \label{sim1}
}

\caption{Visualization of HRG-Net prediction results on a dataset}
\label{grasp}
\end{figure}

We tested our method on two popular public datasets i.e., Cornell Dataset and Jacquard Dataset. The Cornell dataset contains a pair of 885 color maps and depth images with $5110$ positive and $2909$ negative labels labeled by hand. Considering that the Cornell dataset is relatively small and prone to overfitting, we performed a 5-fold cross-validation according to the evaluation criteria from the image perspective (IW) and the object perspective (OW), as in past works \cite{lenz2015deep,kumra2017robotic,redmon2015real}. The input patterns and running times are considered in the comparison. The detailed results are shown in Table \ref{Cornell}. To show the performance of HRG-Net, we visualized part of the images used for validation in the Cornell dataset and the results are shown in Fig.~\ref{grasp} (a) - Fig.~\ref{grasp} (c).
The Jacquard Dataset is a large synthetic dataset containing a total of $11$k objects in $54$k different scenes with a total of 1.1M successfully captured annotations, which is divided into $90\%$ training set and $10\%$ validation set. As shown in Table \ref{Jacquard}, our method performs well in all three channels of RGB, Depth, and RGB-D. 
Also the depth images and RGB images behave closely on the dataset since the background and objects are relatively easy to be distinguished and are highly consistent with the contour boundaries of the objects. 

% It indicates that taking both RGB and depth images as inputs yields improved results. 

\begin{table}[]
\centering
\setlength\tabcolsep{3.5pt}
{
    \footnotesize
    \caption{The accuracy on Cornell grasping dataset. }
    \begin{tabular}{c|c|c|c|c}
        \hline   \multirow{2}{*}{\text { Method }} & \multirow{2}{*}{\text { Input }} & \multicolumn{2}{|c|}{\text { Accuracy(\%) }} & \multirow{2}{*}{\text { Time (ms) }} \\
        \cline { 3 - 4 }  & & \text { IW } & \text { OW } & \\ \hline
        
        Fast Search \cite{jiang2011efficient}& RGB-D & 60.5 &58.3& 5000 \\ \hline
        GG-CNN \cite{morrison2020learning} & D & 73.0&69.0 & \textbf{19}   \\ \hline
        SAE \cite{lenz2015deep} & RGB-D   & 73.9&75.6 &1350 \\ \hline
        ResNet-50x2 \cite{kumra2017robotic}& RGB-D   & 89.2&88.9 & 103 \\ \hline
        AlexNet, MultiGrasp \cite{redmon2015real} & RGB-D &88.0 &87.1 &76 \\ \hline
        STEM-CaRFs\cite{asif2017rgb} & RGB-D   & 88.2&87.5 & - \\ \hline
        GRPN \cite{karaoguz2019object}& RGB   & 88.7&- & 200 \\ \hline
        
        Two-stage closed-loop\cite{wang2016robot} & RGB-D   & 85.3&- &140 \\ \hline
        GraspNet \cite{asif2018graspnet}& RGB-D   & 90.2&90.6 & 24 \\ \hline
        ZF-net \cite{guo2017hybrid}& RGB-D   & 93.2&89.1 & - \\ \hline
        E2E-net \cite{ainetter2021end}  & RGB  & 98.2&- & 63 \\ \hline
        & D   & 93.2&94.3 & \textbf{19}  \\ 
        GR-ConvNet \cite{kumra2020antipodal}& RGB   & 96.6&95.5 & \textbf{19}  \\ 
        & RGB-D   & 97.7&96.6 & 20 \\   \hline  
        & D   & 95.2&94.9 & 41.1 \\ 
        TF-Grasp \cite{wang2022transformer}  & RGB     &96.78   &95.0 & 41.3 \\ 
        & RGB-D   &97.99   &96.7  & 41.6 \\    \hline  \hline
        & \textbf{D}   &\textbf{99.43} &\textbf{96.8} & \textbf{52.6} \\ 
        \textbf{HRG-Net}  & \textbf{RGB}     &\textbf{98.50}   &\textbf{96.7} & \textbf{53.0} \\ 
        & \textbf{RGB-D}   &\textbf{99.50}   &\textbf{97.5}  & \textbf{53.7} \\    \hline

    \end{tabular}
    \label{Cornell}
}
\end{table}
		
\begin{table}[]
    \begin{center}
        \footnotesize
        \caption{The accuracy on Jacquard grasping dataset. } 
        \begin{tabular}{l|l|l|c}
            
            \hline Authors &Method & Input  & Accuracy $(\%)$ \\
            \hline Depierre \cite{depierre2018jacquard} &Jacquard& RGB-D & $74.2$ \\
            Morrison \cite{morrison2020learning}  & GG-CNN2 &D     & 84 \\
            Zhou \cite{zhou2018fully} & FCGN, ResNet-101& RGB & $91.8$ \\
            Alexandre \cite{gariepy2019gq} & GQ-STN &D &$70.8$ \\
            Zhang \cite{zhang2019roi}& ROI-GD &RGB &$90.4$ \\
            \hline
            Stefan \cite{ainetter2021end} &Det Seg &RGB&  $92.59$\\
            Stefan \cite{ainetter2021end} &Det Seg Refine &RGB&  $92.95$\\
            \hline
            Kumra \cite{kumra2020antipodal}     &   GR-ConvNet & D   & 93.7 \\
            Kumra  \cite{kumra2020antipodal}    &   GR-ConvNet & RGB  & 91.8 \\
            Kumra  \cite{kumra2020antipodal}    &   GR-ConvNet & RGB-D   & 94.6 \\
            \hline 
            Wang \cite{wang2022transformer}& TF-Grasp&  D & $93.1$ \\
            Wang \cite{wang2022transformer}& TF-Grasp  &RGB & $93.57$ \\
            Wang \cite{wang2022transformer}& TF-Grasp  &RGB-D & $9 4 . 6$ \\
            \hline & \textbf{HRG-Net} &  \textbf{D} & $\mathbf{95.8}$ \\
            \textbf{Our} & \textbf{HRG-Net}  & \textbf{RGB} & $\mathbf{95.7}$ \\
            & \textbf{HRG-Net}  & \textbf{RGB-D} & $\mathbf{9 6.5}$ \\
            
            \hline
        \end{tabular}
        \label{Jacquard}
    \end{center}
\end{table}

\subsection{Ablation Studies}
	
Our HRG-Net is conceptually different from the traditional design in the robotics community. In contrast to the conventional models that build serial convolutions from high-resolution feature maps to low-resolution maps, our grasping model always maintains high-resolution representation through the parallel branch. To evaluate the role of each component, ablation experiments are carried out on the HRG-Net. Fig.~\ref{grasp} (d) shows the accuracy achieved by the network with different architectures after $10$ epochs of training, where the first one is the Fully Convolutional neural network (FCN) \cite{dai2016r}. Clearly, It achieves an accuracy of about $60\%$ after $10$ epochs. By adding a residual block\cite{he2016deep}, the accuracy is significantly improved because it can enhance gradient propagation and effectively avoid network degradation. It achieves an accuracy of $73\%$ after 10 epochs.
On the other hand, instead of adding a residual block, we use U-Net as a backbone that got similar performance ($75\%$ after 10 epochs) as the FCN with a residual block. Whereas the original CNN suffers from information loss in the feature map due to the use of pooling layers and the fixed receptive field of convolutional kernels. The emergence of residual connections of HRG-Net alleviates this issue and improves the performance of the original CNN. U-Net~\cite{Ronneberger2015} uses skip connections to merge detailed information from low-level layers to high-level layers to enhance model feature fusion and achieve the best results in non-parallel architecture.
However, these three ways still do not leave the architecture of first downsampling and then upsampling, and for problems like grasping, when the object is far from the camera, it often only occupies a few pixel points. Therefore, errors brought by image upsampling are likely to lead to the failure of the grasping task.
For this reason, HRG-Net keeps multiple feature maps of different sizes at the same time, and there exists information interaction between feature maps of different sizes. After testing, the accuracy of HRG-Net can reach $81\%$ with the same training of 10 epochs, which is significantly better than other architectures. This is because such an architecture is able to obtain the rich semantic information in the low-resolution feature maps and maintain the accurate spatial information embedded in the high-resolution feature maps.
		
\begin{table}[]
\begin{center}
\caption{The results for task 2 } 
\begin{tabular}{llc}

\hline Method & Physical  grasp & Success rate $(\%)$ \\
\hline

GG-CNN \cite{Morrison2018} & 81/100 &81\% \\
\hline HRG-Net(Ours)&  93/100  & $93\%$ \\

\hline
\end{tabular}
\label{real_grasp}
\end{center}
\end{table}

% \begin{figure}
%     \centering
%     \includegraphics[width=0.25\textwidth]{Figure_1.pdf} 
%     \caption{Ablation experiments on the Jacquard dataset, the horizontal coordinates from left to right are the original CNN, CNN with residual blocks connection, CNN with skip connection fusion, and HRG-Net, and the vertical coordinates are the accuracy on the validation set after training 10 epochs. {\color{blue}Mingyu: You may need to resize the picture.}}	
%     \label{abl_compare}
% \end{figure}
% \begin{figure}[htbp]
% \centering
% \includegraphics[width=3.5cm]{Figure_1.pdf}
%     \caption{ Ablation experiments on the Jacquard dataset, the horizontal coordinates from left to right are the FCN\cite{dai2016r}, FCN with residual blocks connection\cite{he2016deep}, U-Net, and HRG-Net, and the vertical coordinates are the accuracy on the validation set after training 10 epochs.{\color{blue} Only one picture?}}
% \label{sim1}
% \end{figure}

% \begin{figure}
%     \centering
%     \includegraphics[width=0.50\textwidth]{grasp.pdf} 
%     \caption{The HRG-Net network performance on the Cornell dataset. (a) The visualization of the prediction of grasp bounding box after RGB and depth image are feed into the HRG-Net network. (b) The original RGB images in the dataset and the three outputs of the last layer of the network, which are the grasp quality map $Q$, the grasp angle map $\Phi$, and the grasp width map $W$, respectively.}	
%     \label{grasp}
% \end{figure}
\begin{figure}
    \centering
    \includegraphics[width=0.50\textwidth]{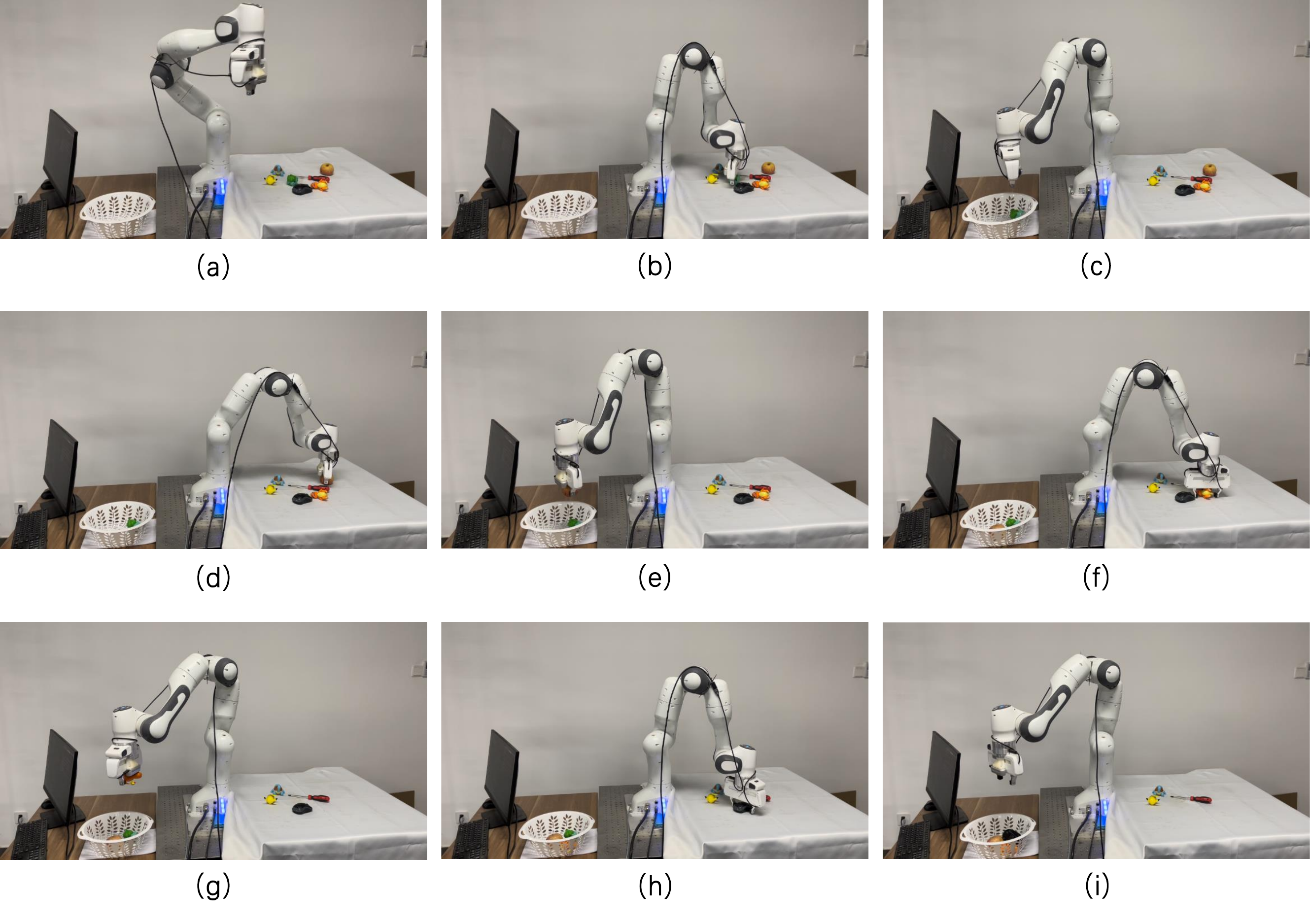} 
    \caption{Demonstration of HRG-Net in multi-object grasping }	
    \label{mograsp}
\end{figure}

\begin{figure*}
    \centering
    \includegraphics[width=1\textwidth]{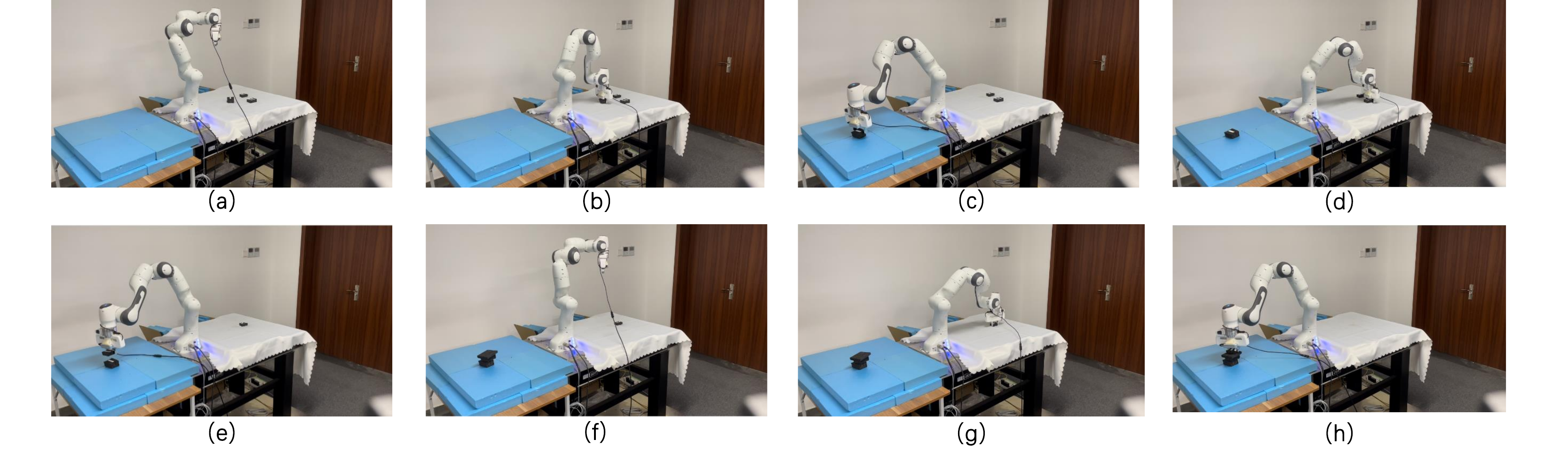} 
    \caption{HGR-Net for a human-computer collaboration stack task. }	
    \label{stacking}
\end{figure*}

\begin{figure}
    \centering
    \includegraphics[width=0.40\textwidth]{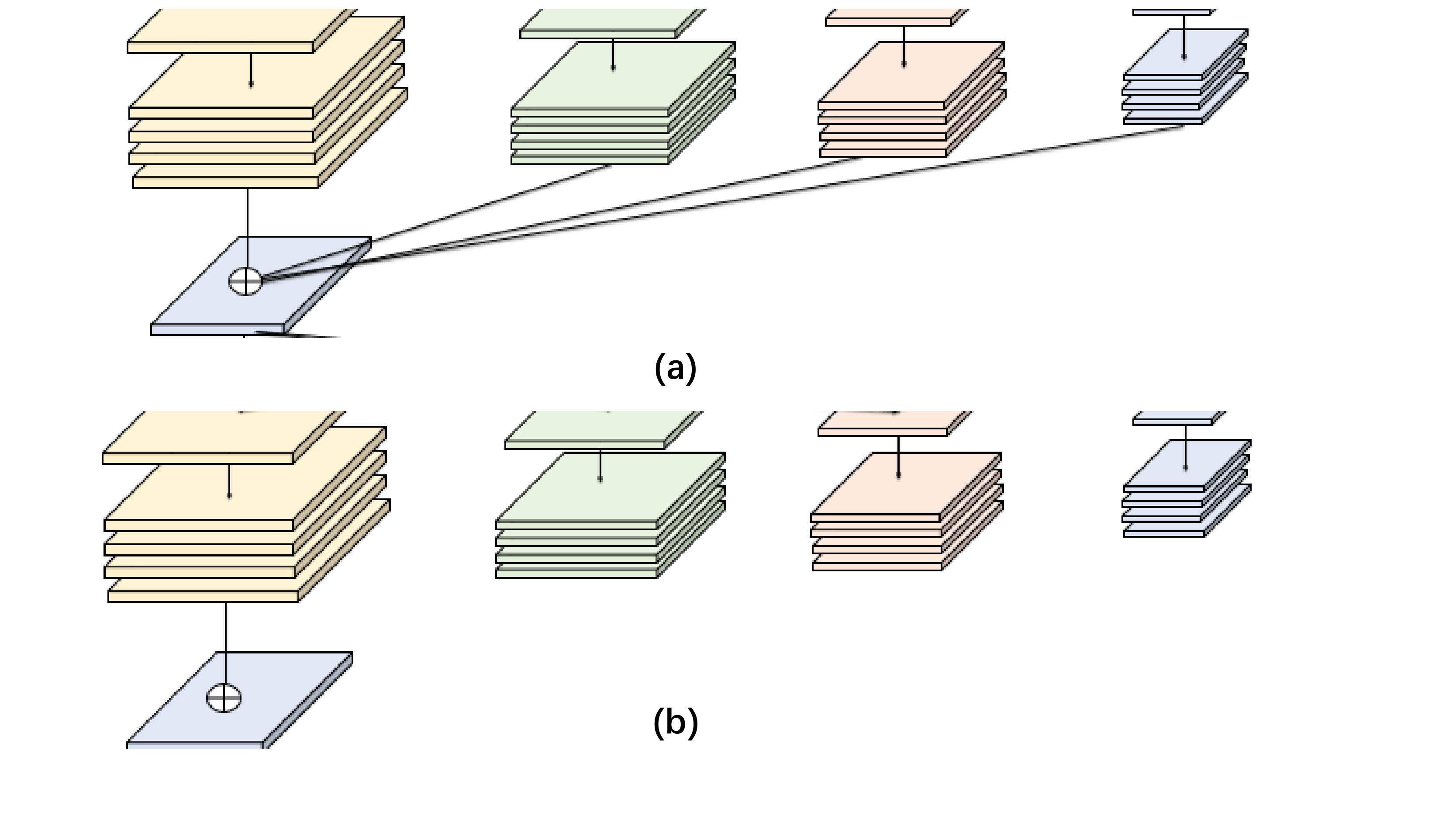} 
    \caption{The last two layers of Fig. \ref{network} (a) Only output the representation from the high-resolution convolution stream. (b) Concatenate the (upsampled)
        representations that are from all the resolutions }	
    \label{last_fusion}
\end{figure}

The last feature information is fused in the following two ways to generate relevant grasp representations. As in Fig.~\ref{last_fusion} (a), features from all branches at different resolutions are combined to predict the corresponding grasp configurations.  In Fig.~\ref{last_fusion} (b), only the feature maps from the highest resolution branch are used to generate the final grasp representations. The performances of the two approaches over the Cornell and Jacquard dataset are shown in Table \ref{layer-fusion}. We can see that the final output of the feature map with a mix of sizes gives a better result 
then the feature map considering only the highest resolution.

\begin{table}[]
\centering
{
    \footnotesize
    \caption{Comparison between using and not using Layer-fusion }
    \begin{tabular}{c|c|c}
        \hline
        \multicolumn{3}{c}{The accuracy on Cornell Grasping Results} \\ \hline
        & With Layer-fusion & Without Layer-fusion  \\ \hline
        RGB & $98.50\%$& $97.50\%$\\ \hline
        Depth & $99.43\%$& $98.00\%$\\ \hline
        RGB+Depth & $99.50\%$& $98.50\%$\\ \hline
        \multicolumn{3}{c}{The accuracy on Jacquard Grasping Results} \\ \hline
        & With Layer-fusion & Without Layer-fusion  \\ \hline
        RGB & $95.70\%$& $94.42\%$\\ \hline
        Depth & $95.80\%$& $95.26\%$\\ \hline
        RGB+Depth & $96.50\%$& $95.82\%$\\ \hline

        %architecture &
        %  \multicolumn{4}{c|}{Encoder} &
        %  \\ \hline
    \end{tabular}
    \label{layer-fusion}
}
\end{table}

\subsection{Experiments in real physical environments}

To test the generalization capability of unseen objects, we selected novel objects that are not available in the dataset. We will design three sets of experiments for objects that do not appear in these datasets, with the aim of highlighting the advantages of our algorithm.

In the first task, the robot manipulation needs to move all the objects on the table to the basket, the main challenge of this task is that the objects in the workspace are far away from the camera and may stick together, so it is very easy to receive interruptions from other objects in the process of grasping e.g., Fig.~\ref{failed}. In this experiment, we compared Chu\cite{chu2018real} et al. and GG-CNN\cite{Morrison2018}. We found that Chu's method took about $9$ minutes and $15$ seconds to complete the task, GG-CNN\cite{Morrison2018} took 8 minutes and 20 seconds to complete the task, and ours only needed 4 minutes and 13 seconds to complete the task. The main reason is the first two lose spatial accuracy due to downsampling, resulting in poor localization, and frequent interruptions by other objects adhering to the target object. In contrast, our model reduced the risk of imprecision of spatial information via a high-resolution feature map and downsampling. It is also found that several objects with sophisticated shapes are challenging to grasp in practice. For instance, the toy depicted in Fig.~\ref{mograsp} has frequently bumped the gripper because of its tail.

In the second task, we consider the adaptability of different networks' architecture to unseen objects, and compared with GG-CNN~\cite{Morrison2018}. In our experiments, we tested the robustness of both methods to objects with random initial positions and dynamic objects (here the distance is less than $10$cm) and found that both methods can have some adaptability, but our HRG-Net performs better. This can be seen in table~\ref{real_grasp} where we tested each method $100$ times, $50$ times for random positions, and $50$ times for dynamic objects.

% We tested each method 100 times, 50 times for random positions, and 50 times for dynamic objects, and the results of the tests are shown in the table~\ref{real_grasp}.

The last task is to build a block building, which requires robot manipulation to complete the grasping and placing the blocks in a specific order. The whole process needs to ensure that the block  building does not fall down. The difficulty of this task is how to ensure accurate positioning and motion planning. 
Because our motion planning algorithm online predicts the optimal position and pose during execution, and the size and position of objects from different viewpoints can vary greatly. The resulting error due to upsampling in GG-CNN~\cite{Morrison2018} will be more obvious, which is reflected in the experiment that the blocks cannot be placed correctly. 
In contrast, HRG-Net performs well due to its capability of high-resolutions.
Due to space limitation, we only include the screenshots of the experiments using HRG-Net in Fig.~\ref{stacking}, and the detailed comparisons of performances can be found in the webpage: \url{https://github.com/USTCzzl/HRG_Net}.

\section{Conclusion}

In this paper, we propose a framework of robotic visual grasping that maintains high-resolution representations and connects the high-to-low resolution convolution streams simultaneously.
Our design is effective in boosting the performance of perception tasks by using parallel branches and fusing information between different resolutions rather than a single branch stacked convolutional layers. Physical experimental results including comparisons with the mainstream baselines show that HRG-Net delivers better performance for a large margin on several datasets. In conclusion, accurate spatial information integrated with motion planning plays an important role as rich semantic information, and HRG-Net can better address problems of the far distance between the initial camera and target. Future work will further explore cases where grasping objects are too small or the depth of information is not obvious.

% \vfill
%  Future work will further account cases where grasping objects are too small or the depth information is not obvious.
% This paper revisits the design paradigm of convolutional neural networks in robotic perception tasks. Our model maintains high-resolution representations and connects the high-to-low resolution convolution streams simultaneously. We demonstrate that the design using parallel branches and fusing information between different resolutions rather than a single branch stacked convolutional layers is more effective in boosting the performance of robotic perception tasks. The experimental results including comparisons with the mainstream open source algorithms in a physical environment show that the high-resolution model delivers better performance and outperforms baselines by a large margin on several datasets. In summary,  for perception-based robotic grasping, we can conclude that accurate spatial information integrated with motion planning plays an important role as rich semantic information, and HRG-Net can better solve the problem of far distance between the initial camera and target.
%  Future work will further account cases where grasping objects are too small or the depth information is not obvious.
    
%\end{thebibliography}
\bibliographystyle{IEEEtran}
\bibliography{reference.bib}

\begin{thebibliography}{10}
\providecommand{\url}[1]{#1}
\csname url@samestyle\endcsname
\providecommand{\newblock}{\relax}
\providecommand{\bibinfo}[2]{#2}
\providecommand{\BIBentrySTDinterwordspacing}{\spaceskip=0pt\relax}
\providecommand{\BIBentryALTinterwordstretchfactor}{4}
\providecommand{\BIBentryALTinterwordspacing}{\spaceskip=\fontdimen2\font plus
\BIBentryALTinterwordstretchfactor\fontdimen3\font minus
  \fontdimen4\font\relax}
\providecommand{\BIBforeignlanguage}[2]{{%
\expandafter\ifx\csname l@#1\endcsname\relax
\typeout{** WARNING: IEEEtran.bst: No hyphenation pattern has been}%
\typeout{** loaded for the language `#1'. Using the pattern for}%
\typeout{** the default language instead.}%
\else
\language=\csname l@#1\endcsname
\fi
#2}}
\providecommand{\BIBdecl}{\relax}
\BIBdecl

\bibitem{DBLP:conf/icra/MorrisonCL19}
\BIBentryALTinterwordspacing
D.~Morrison, ``Multi-view picking: Next-best-view reaching for improved
  grasping in clutter,'' in \emph{International Conference on Robotics and
  Automation, {ICRA} 2019, Montreal, QC, Canada, May 20-24, 2019}.\hskip 1em
  plus 0.5em minus 0.4em\relax {IEEE}, 2019, pp. 8762--8768. [Online].
  Available: \url{https://doi.org/10.1109/ICRA.2019.8793805}
\BIBentrySTDinterwordspacing

\bibitem{lenz2015deep}
I.~Lenz, H.~Lee, and A.~Saxena, ``Deep learning for detecting robotic grasps,''
  \emph{Int. J. Robotics Res.}, vol.~34, no. 4-5, pp. 705--724, 2015.

\bibitem{kumra2020antipodal}
S.~Kumra, S.~Joshi, and F.~Sahin, ``Antipodal robotic grasping using generative
  residual convolutional neural network,'' in \emph{2020 IEEE/RSJ Int. Conf.
  Intel. Robot. Syst.}\hskip 1em plus 0.5em minus 0.4em\relax IEEE, 2020, pp.
  9626--9633.

\bibitem{ainetter2021end}
S.~Ainetter and F.~Fraundorfer, ``End-to-end trainable deep neural network for
  robotic grasp detection and semantic segmentation from rgb,'' in \emph{Proc.
  IEEE Int. Conf. Robot.Automat.}\hskip 1em plus 0.5em minus 0.4em\relax IEEE,
  2021, pp. 13\,452--13\,458.

\bibitem{kumra2017robotic}
S.~Kumra and C.~Kanan, ``Robotic grasp detection using deep convolutional
  neural networks,'' in \emph{Proc. IEEE/RSJ Int. Conf. Intell. Robots Syst},
  2017, pp. 769--776.

\bibitem{Redmon2015}
J.~Redmon and A.~Angelova, ``Real-time grasp detection using convolutional
  neural networks,'' in \emph{IEEE Proc. Int. Conf. Robot. Autom.}\hskip 1em
  plus 0.5em minus 0.4em\relax IEEE, 2015, pp. 1316--1322.

\bibitem{wang2016robot}
Z.~Wang, Z.~Li, B.~Wang, and H.~Liu, ``Robot grasp detection using multimodal
  deep convolutional neural networks,'' \emph{Advances in Mechanical
  Engineering}, vol.~8, no.~9, p. 1687814016668077, 2016.

\bibitem{lecun1998gradient}
Y.~LeCun, L.~Bottou, Y.~Bengio, and P.~Haffner, ``Gradient-based learning
  applied to document recognition,'' \emph{Proceedings of the IEEE}, vol.~86,
  no.~11, pp. 2278--2324, 1998.

\bibitem{krizhevsky2017imagenet}
A.~Krizhevsky, I.~Sutskever, and G.~E. Hinton, ``Imagenet classification with
  deep convolutional neural networks,'' \emph{Communications of the ACM},
  vol.~60, no.~6, pp. 84--90, 2017.

\bibitem{wang2015places205}
L.~Wang, S.~Guo, W.~Huang, and Y.~Qiao, ``Places205-vggnet models for scene
  recognition,'' \emph{arXiv preprint arXiv:1508.01667}, 2015.

\bibitem{he2016deep}
K.~He, X.~Zhang, S.~Ren, and J.~Sun, ``Deep residual learning for image
  recognition,'' in \emph{Proceedings of the IEEE conference on computer vision
  and pattern recognition}, 2016, pp. 770--778.

\bibitem{Ronneberger2015}
O.~Ronneberger, P.~Fischer, and T.~Brox, ``U-net: Convolutional networks for
  biomedical image segmentation,'' in \emph{Int. Conf. Med. Image Comput.
  Computer-assisted Interv.}\hskip 1em plus 0.5em minus 0.4em\relax Springer,
  2015, pp. 234--241.

\bibitem{Lenz2015}
I.~Lenz, H.~Lee, and A.~Saxena, ``Deep learning for detecting robotic grasps,''
  \emph{Int. J. Robot. Res.}, vol.~34, no. 4-5, pp. 705--724, 2015.

\bibitem{depierre2018jacquard}
A.~Depierre, E.~Dellandr{\'e}a, and L.~Chen, ``Jacquard: A large scale dataset
  for robotic grasp detection,'' in \emph{Proc. IEEE/RSJ Int. Conf. Intell.
  Robots Syst.}, 2018, pp. 3511--3516.

\bibitem{ramisa2014learning}
A.~Ramisa, G.~Alenya, F.~Moreno-Noguer, and C.~Torras, ``Learning rgb-d
  descriptors of garment parts for informed robot grasping,'' \emph{Engineering
  Applications of Artificial Intelligence}, vol.~35, pp. 246--258, 2014.

\bibitem{jiang2011}
\emph{{IEEE} International Conference on Robotics and Automation, {ICRA} 2011,
  Shanghai, China, 9-13 May 2011}.\hskip 1em plus 0.5em minus 0.4em\relax
  {IEEE}, 2011.

\bibitem{DBLP:journals/ral/ShaoFJNLSOKB20}
\BIBentryALTinterwordspacing
L.~Shao, F.~Ferreira, M.~Jorda, V.~Nambiar, J.~Luo, E.~Solowjow, J.~A. Ojea,
  O.~Khatib, and J.~Bohg, ``Unigrasp: Learning a unified model to grasp with
  multifingered robotic hands,'' \emph{{IEEE} Robotics Autom. Lett.}, vol.~5,
  no.~2, pp. 2286--2293, 2020. [Online]. Available:
  \url{https://doi.org/10.1109/LRA.2020.2969946}
\BIBentrySTDinterwordspacing

\bibitem{zhang2019roi}
H.~Zhang, X.~Lan, S.~Bai, X.~Zhou, Z.~Tian, and N.~Zheng, ``Roi-based robotic
  grasp detection for object overlapping scenes,'' in \emph{Proc. IEEE Int.
  Conf. Intell. Robots Syst.}, 2019, pp. 4768--4775.

\bibitem{morrison2020learning}
D.~Morrison, P.~Corke, and J.~Leitner, ``Learning robust, real-time, reactive
  robotic grasping,'' \emph{Int. J. Robotics Res.}, vol.~39, no. 2-3, pp.
  183--201, 2020.

\bibitem{cao2021lightweight}
H.~Cao, G.~Chen, Z.~Li, J.~Lin, and A.~Knoll, ``Lightweight convolutional
  neural network with gaussian-based grasping representation for robotic
  grasping detection,'' \emph{arXiv preprint arXiv:2101.10226}, 2021.

\bibitem{ren2015faster}
S.~Ren, K.~He, R.~Girshick, and J.~Sun, ``Faster r-cnn: Towards real-time
  object detection with region proposal networks,'' \emph{Advances in neural
  information processing systems}, vol.~28, 2015.

\bibitem{DBLP:conf/icarm/LuoTJPX20}
\BIBentryALTinterwordspacing
Z.~Luo, B.~Tang, S.~Jiang, M.~Pang, and K.~Xiang, ``Grasp detection based on
  faster region {CNN},'' in \emph{5th International Conference on Advanced
  Robotics and Mechatronics, {ICARM} 2020, Shenzhen, China, December 18-21,
  2020}.\hskip 1em plus 0.5em minus 0.4em\relax {IEEE}, 2020, pp. 323--328.
  [Online]. Available: \url{https://doi.org/10.1109/ICARM49381.2020.9195274}
\BIBentrySTDinterwordspacing

\bibitem{DBLP:conf/icra/PintoG16}
\BIBentryALTinterwordspacing
L.~Pinto and A.~Gupta, ``Supersizing self-supervision: Learning to grasp from
  50k tries and 700 robot hours,'' in \emph{2016 {IEEE} International
  Conference on Robotics and Automation, {ICRA} 2016, Stockholm, Sweden, May
  16-21, 2016}, D.~Kragic, A.~Bicchi, and A.~D. Luca, Eds.\hskip 1em plus 0.5em
  minus 0.4em\relax {IEEE}, 2016, pp. 3406--3413. [Online]. Available:
  \url{https://doi.org/10.1109/ICRA.2016.7487517}
\BIBentrySTDinterwordspacing

\bibitem{wang2022transformer}
S.~Wang, Z.~Zhou, and Z.~Kan, ``When transformer meets robotic grasping:
  Exploits context for efficient grasp detection,'' \emph{IEEE Robot. Autom.
  Lett.}, vol.~7, no.~3, pp. 8170--8177, 2022.

\bibitem{Morrison2018}
D.~Morrison, P.~Corke, and J.~Leitner, ``Closing the loop for robotic grasping:
  A real-time, generative grasp synthesis approach,'' \emph{arXiv preprint
  arXiv:1804.05172}, 2018.

\bibitem{he2017mask}
K.~He, G.~Gkioxari, P.~Doll{\'a}r, and R.~Girshick, ``Mask r-cnn,'' in
  \emph{Proceedings of the IEEE international conference on computer vision},
  2017, pp. 2961--2969.

\bibitem{sun2019deep}
K.~Sun, B.~Xiao, D.~Liu, and J.~Wang, ``Deep high-resolution representation
  learning for human pose estimation,'' in \emph{Proceedings of the IEEE/CVF
  conference on computer vision and pattern recognition}, 2019, pp. 5693--5703.

\bibitem{wang2020deep}
J.~Wang, K.~Sun, T.~Cheng, B.~Jiang, C.~Deng, Y.~Zhao, D.~Liu, Y.~Mu, M.~Tan,
  X.~Wang \emph{et~al.}, ``Deep high-resolution representation learning for
  visual recognition,'' \emph{IEEE Trans. Pattern Anal. Mach. Intell.},
  vol.~43, no.~10, pp. 3349--3364, 2020.

\bibitem{morrison2019multi}
D.~Morrison, P.~Corke, and J.~Leitner, ``Multi-view picking: Next-best-view
  reaching for improved grasping in clutter,'' in \emph{2019 International
  Conference on Robotics and Automation (ICRA)}.\hskip 1em plus 0.5em minus
  0.4em\relax IEEE, 2019, pp. 8762--8768.

\bibitem{DBLP:conf/iclr/LoshchilovH19}
\BIBentryALTinterwordspacing
I.~Loshchilov and F.~Hutter, ``Decoupled weight decay regularization,'' in
  \emph{7th International Conference on Learning Representations, {ICLR} 2019,
  New Orleans, LA, USA, May 6-9, 2019}, 2019. [Online]. Available:
  \url{https://openreview.net/forum?id=Bkg6RiCqY7}
\BIBentrySTDinterwordspacing

\bibitem{redmon2015real}
J.~Redmon and A.~Angelova, ``Real-time grasp detection using convolutional
  neural networks,'' in \emph{Proc. IEEE Int. Conf. Robot. Autom.}, 2015, pp.
  1316--1322.

\bibitem{jiang2011efficient}
Y.~Jiang, S.~Moseson, and A.~Saxena, ``Efficient grasping from rgbd images:
  Learning using a new rectangle representation,'' in \emph{Proc. IEEE Int.
  Conf. Robot. Automat.}, 2011, pp. 3304--3311.

\bibitem{asif2017rgb}
U.~Asif, M.~Bennamoun, and F.~A. Sohel, ``Rgb-d object recognition and grasp
  detection using hierarchical cascaded forests,'' \emph{IEEE Trans. on
  Robotics}, vol.~33, no.~3, pp. 547--564, 2017.

\bibitem{karaoguz2019object}
H.~Karaoguz and P.~Jensfelt, ``Object detection approach for robot grasp
  detection,'' in \emph{Proc. IEEE Int. Conf. Robot.Automat.}, 2019, pp.
  4953--4959.

\bibitem{asif2018graspnet}
U.~Asif, J.~Tang, and S.~Harrer, ``Graspnet: An efficient convolutional neural
  network for real-time grasp detection for low-powered devices.'' in
  \emph{IJCAI}, vol.~7, 2018, pp. 4875--4882.

\bibitem{guo2017hybrid}
D.~Guo, F.~Sun, H.~Liu, T.~Kong, B.~Fang, and N.~Xi, ``A hybrid deep
  architecture for robotic grasp detection,'' in \emph{Proc. IEEE Int. Conf.
  Robot.Automat.}, 2017, pp. 1609--1614.

\bibitem{zhou2018fully}
X.~Zhou, X.~Lan, H.~Zhang, Z.~Tian, Y.~Zhang, and N.~Zheng, ``Fully
  convolutional grasp detection network with oriented anchor box,'' in
  \emph{Proc. IEEE Int. Conf. Intell. Robots Syst.}, 2018, pp. 7223--7230.

\bibitem{gariepy2019gq}
A.~Gari{\'e}py, J.-C. Ruel, B.~Chaib-Draa, and P.~Giguere, ``Gq-stn: Optimizing
  one-shot grasp detection based on robustness classifier,'' in \emph{Proc.
  IEEE/RSJ Int. Conf. Intell. Robots Syst.}, 2019, pp. 3996--4003.

\bibitem{dai2016r}
J.~Dai, Y.~Li, K.~He, and J.~Sun, ``R-fcn: Object detection via region-based
  fully convolutional networks,'' \emph{Advances in neural information
  processing systems}, vol.~29, 2016.

\bibitem{chu2018real}
F.-J. Chu, R.~Xu, and P.~A. Vela, ``Real-world multiobject, multigrasp
  detection,'' \emph{{IEEE} Robotics Autom. Lett.}, vol.~3, no.~4, pp.
  3355--3362, 2018.

\end{thebibliography}

    %\newpage

    
    %\vfill
    

\end{document}